**Improving the Accuracy of Freight Mode Choice Models: A Case Study Using the 2017 CFS PUF Data Set and Ensemble Learning Techniques**[1]

*Corresponding author: Diyi Liu
**Diyi Liu**
Department of Civil and Environmental Engineering
University of Tennessee, Knoxville, Tennessee, USA
Email: dliu27@vols.utk.edu

**Hyeonsup Lim**
Buildings and Transportation Science Division
Oak Ridge National Laboratory, Oak Ridge, Tennessee, USA
Email: limh@ornl.gov

**Majbah Uddin**
Buildings and Transportation Science Division
Oak Ridge National Laboratory, Oak Ridge, Tennessee, USA
Email: uddinm@ornl.gov

**Yuandong Liu**
Buildings and Transportation Science Division
Oak Ridge National Laboratory, Oak Ridge, Tennessee, USA
Email: yuandong_liu@hotmail.com

**Lee D. Han**
Department of Civil and Environmental Engineering
University of Tennessee, Knoxville, Tennessee, USA
Email: lhan@utk.edu

**Ho-ling Hwang**
Buildings and Transportation Science Division
Oak Ridge National Laboratory, Oak Ridge, Tennessee, USA
Email: hwanghlc@gmail.com

**Shih-Miao Chin**
Buildings and Transportation Science Division
Oak Ridge National Laboratory, Oak Ridge, Tennessee, USA
Email: chinsm@comcast.net

Word Count: 11,290 words
*Submitted [Oct 6, 2023]*

---

[1] Notice: This manuscript has been authored by UT-Battelle, LLC, under contract DE-AC05-00OR22725 with the US Department of Energy (DOE). The US government retains and the publisher, by accepting the article for publication, acknowledges that the US government retains a nonexclusive, paid-up, irrevocable, worldwide license to publish or reproduce the published form of this manuscript, or allow others to do so, for US government purposes. DOE will provide public access to these results of federally sponsored research in accordance with the DOE Public Access Plan.

**Abstract**

The US Census Bureau has collected two rounds of experimental data from the Commodity Flow Survey, providing shipment-level characteristics of nationwide commodity movements, published in 2012 (i.e., Public Use Microdata) and in 2017 (i.e., Public Use File). With this information, data-driven methods have become increasingly valuable for understanding detailed patterns in freight logistics. In this study, we used the 2017 Commodity Flow Survey Public Use File data set to explore building a high-performance freight mode choice model, considering three main improvements: (1) constructing local models for each separate commodity/industry category; (2) extracting useful geographical features, particularly the derived distance of each freight mode between origin/destination zones; and (3) applying additional ensemble learning methods such as stacking or voting to combine results from local and unified models for improved performance. The proposed method achieved over 92% accuracy without incorporating external information, an over 19% increase compared to directly fitting Random Forests models over 10,000 samples. Furthermore, SHAP (Shapely Additive Explanations) values were computed to explain the outputs and major patterns obtained from the proposed model. The model framework could enhance the performance and interpretability of existing freight mode choice models.

# 1    Introduction

Understanding national-level freight mode choice is a vital research area regarding sustainability, planning, policymaking, and economic analysis. The share of freight modes fluctuates over time because of the ever-changing operational characteristics and environment. Therefore, a comprehensive grasp of the underlying determinants that influence mode choice is crucial for optimizing freight transportation systems.

Recently, the landscape of freight logistics has experienced rapid changes under the influence of multiple factors, including technological advancements, infrastructure deterioration, labor shortages, and the fast growth of electronic commerce. For instance, the adoption of autonomous and electric vehicles holds great potential in reducing truck operating expenses (Monios & Bergqvist, 2019), leading to a possible shift from rail to road transport for certain cargo types. The surge in online shopping has placed immense strain on parcel delivery services worldwide (Zenezini et al., 2018), particularly when freight capacity is constrained by factors such as a shortage of truck drivers (Burks & Monaco, 2019; Hobbs, 2020). Furthermore, new operational practices in rail systems such as shift scheduling have been linked to fatigue, which could result in accidents (Rudin-Brown et al., 2019; Zhang et al., 2021). Mitigating these challenges requires a thorough understanding of the flow of goods across various transportation modes.

The Commodity Flow Survey (CFS), a collaborative effort of the US Census Bureau and the Bureau of Transportation Statistics, provides valuable insights into the characteristics of freight shipments in the United States. This shipper-based survey has been conducted every 5 years since the early 1990s and covers a broad spectrum of industries, encompassing approximately 100,000 establishments. The CFS data have been widely used by various federal, state, and local and regional agencies to understand freight commodity flows, evaluate transportation demand and infrastructures, and facilitate informed decision-making processes.

The CFS provides valuable data for analyzing freight transportation mode choices. Furthermore, experimental data products—Public Use Microdata in 2012 and Public Use File (PUF) in 2017—have allowed for more detailed analyses at the shipment level (Arencibia et al., 2015; Larranaga et al., 2017; Samimi et al., 2011). These surveys present a unique opportunity to employ a data-oriented, disaggregate-level approach to better understand freight mode selection.

This study focuses solely on the 2017 CFS PUF data and proposes a framework to enhance model performance by considering three key improvements: (1) constructing local models for each individual commodity/industry category; (2) extracting useful geographical features, particularly the derived distance of each freight mode between origin and destination zones; and (3) implementing additional ensemble learning methods, such as stacking or voting, to combine results from local and unified models for superior performance.

This article is organized as follows. In **Section 2**, prior practices of fitting machine learning models for both general classification tasks and transportation/freight mode choice tasks are discussed. **Section 3** provides a definition of the mode classification used in this study, along with a summary of the CFS PUF data set. **Section 4** describes the data processing steps, the model selection procedures, and the machine learning algorithms adopted. **Section 5** presents model results through both performance metrics and a receiver operating characteristic curve.

The model is also explained using different model interpretation methods including variable importance, SHAP values, etc. Finally, **Section 6** concludes with an overview of lessons learned from this study, limitations, and future research directions.

## 2  Literature Review

The complexity of the freight system, with its multiple modes of transportation, operators, and routes, makes it difficult to obtain all operational data directly from the field. Consequently, surveys are one of the most viable methods for understanding freight choices for most applications. Two types of surveys are commonly used: stated preference (SP) and revealed preference (RP). SP surveys allow responders to indicate their preferred choice given specific scenarios and choices, whereas RP surveys generate reports that reflect actual choices made for their existing operations. SP surveys are often designed to gain insights into adopting new technologies, infrastructures, and policies. For example, Larranaga et al. (2017) designed an SP survey for six scenarios targeting policies that can encourage intramodality in Rio Grande do Sul, Brazil. RP surveys have become increasingly popular in recent years because of the lowered cost in collecting large data sets. Rich et al. (2009) in Denmark used an RP survey, the FEMEX/COMVIC RP data set, to build a multinomial logistic regression (MNL) model considering the effects of free flow time, congestion time, and waiting time for each traffic mode.

In the United States, numerous SP surveys have been established to understand freight choices within specific regions, industries, and commodities. For example, Samimi et al. (2011) built a behavior model of freight choice based on a survey conducted for 316 firms. Conducting such surveys incurs high labor and monetary costs. The recent establishment of the CFS PUF data set, as an RP data set, has presented an opportunity for researchers to access millions of freight shipment records. This public data set provides a nationwide scope for understanding freight choice in the United States using millions of records revealing freight shipments. Two shipment-level microdata sets were released for 2012 and 2017. The CFS PUF data set contains a wide range of industry and commodity types, as described in **Section 3**. For the 2017 survey, approximately 60,000 establishments were involved, and 5,987,535 shipments were recorded (2017 Commodity Flow Survey (CFS) Public Use File (PUF) Data Users Guide, 2020). With such scope and complexity, data-oriented methods such as machine learning may be preferred over traditional methods to attain high prediction performance in accuracies.

In the field of behavior modeling, two primary approaches are used: regression-based methods and machine learning methods. The regression-based methods, such as MNL, have proven useful in identifying statistical significance surveying transportation mode choice (Arencibia et al., 2015; Holguín-Veras et al., 2021; Keya et al., 2018; Uddin et al., 2021). For models with only two classes, MNL reduces to binary logit regression (LR). Hierarchical classification structures, such as those found in freight modes where intermodal modes like water-truck mode and rail-truck mode fall under intermodal mode, can be modeled with nested logistic regression (NL), which nests multiple MNLs together. These three models (i.e., MNL, LR, and NL) are known as *random utility models* because they assume an individual's preference can be described by a utility function. This approach is commonly employed in traditional freight choice modeling (Arencibia et al., 2015; Holguín-Veras et al., 2021; Larranaga et al., 2017; Moschovou & Giannopoulos, 2012; Norojono & Young, 2003; Rich et al., 2009; Samimi et al., 2011; Shen &

Wang, 2012; Uddin et al., 2021; Wang et al., 2013). These models are augmented by various explanatory variables that influence the utilities or attractiveness of each choice. Sensitivity analysis can be used to evaluate the effect of different influencing factors (Arencibia et al., 2015; Larranaga et al., 2017; Samimi et al., 2011). Another method, random regret minimization, employs a similar structure but uses different formulations that focus on the compromise effect of alternatives for modeling freight mode choice using the CFS PUF data set (Keya et al., 2018, 2019). Although these parametric models are valuable for analyses such as sensitivity analysis and econometric analysis, the lack of flexibilities of traditional models often limit their overall performance when applied on millions of data records.

For the classification of CFS PUF, which has thousands of subcategories and millions of records, machine learning methods may be preferred over traditional methods for their superior performance. To evaluate the effectiveness of various machine learning approaches, such as Naïve Bayes (NB), support vector machine (SVM), artificial neural network, *k*-nearest neighbors (KNN), decision tree (DT), random forest (RF), boosting DT, and bagging DT, Uddin et al. (2021) conducted a series of experiments with different numbers of training samples. Although the performance of DT and RF methods reached around 75.4% accuracy, which is significantly higher than utility-based MNL performance of 42.0% accuracy, further improvements can be made using ensemble techniques. Moreover, Uddin et al. (2021) and Keya et al. (2019) augmented the CFS Public Use Microdata data set by adding several secondary information such as local weather, logistics information, and origin and destination facility densities, which could limit the applicability of these models to public users. Models with accuracies below 80% may still not fully exploit the provided shipment information in terms of predicting freight mode. To further enhance model performance, other machine learning practices must be adopted that utilize big data from other disciplines.

The topic of transportation mode choice has been the focus of recent studies recently. A comprehensive synthesis of these studies during the last decade is presented in **Table 1**. Notably, with the exception of Uddin et al. (2021), all other studies have primarily focused on individual trip mode rather than freight mode choice, and they are arranged in ascending chronological order of publication. The observations from these studies are as follows: 1) Conventional methods such as MNL/LR do not perform well; 2) Machine learning methods such as NB, SVM, DT, MLP show better performance; 3) Tree-based models like RF and boosting decision trees (BOOST) are likely to achieve higher accuracies, probably due to the tabular nature of mode choice data, as indicated by these comparison studies. However, a limitation inherent in these methodologies is their tendency to prioritize the identification of fundamental input variables, often neglecting the derived explanatory features. Additionally, machine learning offers other advanced techniques to further enhance model's performance.

**Table 1** Summary of recent comparative study comparing between machine learning algorithms and traditional methods in the application of transportation mode choice.

| Paper | Data sources & sizes | Study scope | Mode | Method tested | Training and testing sets | Best performed model reported |
|---|---|---|---|---|---|---|
| Uddin et al. (2021) | 2012 CFS data; 4,547,661 shipments from approximately 60,000 responding establishments; | United States | For-hire truck, private truck, parcel service, air, other mode | MNL; NB; SVM; ANN; CART; RF; BOOST; KNN | Performance tested under different sample sizes (136,073-2,267,842); tested training set (10%-60%). | Among all investigated classifiers, the highest model, a RF model, reached about 82% accuracy. |
| Omrani (2015) | Luxemburg PSELL survey; 9,500 individuals | Luxemburg | Car, PT, walk/bike | MNL; ANN-MLP; ANN-BLF; SVM | 60% training set; 40% testing set | Accuracy reported: MNL (0.65); SVM (0.68); ANN-RBF (0.80); ANN-MLP (0.82) |
| Nam et al. (2017) | Switzerland mode choice survey for long-distance travel; 6,768 individual observations | Switzerland | Car, rail, Metro | NL; CNL; ANN; ANN-MLP; DNN | Best outcomes correspond to 70% training set; 30% testing set | Accuracy reported: NL (0.64); CNL (0.64); ANN-MLP (0.63); DNN (0.67) |
| Hagenauer & Helbich (2017) | Dutch National Travel Survey (NTS); 230,608 trips reported from 69,918 individuals | Netherlands | car, walk, bike, PT | MNL; NB; SVM; ANN; RF; BOOST; BAG | 69,918 individuals with 230,608 trips reported | MNL (0.561), ANN (0.606), NB (0.602), BOOST (0.801), SVM (0.825), BAG (0.906), RF (0.914) |
| Cheng et al. (2019) | Household survey; | Nanjing, China | Walk, bike, EM, PT, car | MNL; RF; SVM; BOOST | 80% training set; 20% testing set | RF and SVM provides the best performance |
| Zhao et al. (2020) | State Preference Survey at University of Michigan, Ann Arbor campus; 8,141 observations from 1,163 individuals | Ann Arbor, Michigan, United States | car, walk, bike, PT | MNL; NB; NN; CART; SVM; RF; BOOST; BAG | 10-fold cross validation for tunning. | MNL(0.647); mixed logit (0.631); RF (0.856) |
| Kim (2021) | 2016 NHTS. 172,889 trips | Seoul, Korea | Car, bike, PT, walk | ANN, RF, XGBoost | 75% for training 25% for testing | ANN (0.882); RF (0.909); XGBoost (0.923) |
| Tamim Kashifi et al. (2022) | Dutch national travel survey (NTS); 230,608 trips reported from 69,918 individuals | Netherlands | car, walk, bike, PT | LR; MLP; DT; RF; LightGBDT | 10% testing set & 90% training set; Ten-fold cross validation for tunning; | Reported accuracies: LR (0.58), MLP (0.62), DT (0.80), RF (0.87), LightGBDT (0.89) |

Notes for abbreviation and acronyms: LR: Logic Regression; MNL: Multinominal Logistic Regression; NL: Nested Logit Model; CNL: Cross Nested Logit Model; NB: Naïve Bayes; SVM: Support Vector Machine; CART: Classification and Regression Tree; RF: Random Forest; BOOST: Gradient Boosting Decision Tree; BAG: Bagging Decision Tree; NN: Neural Networks; ANN: Artificial Neural Networks; MLP: Multi-layer Perceptron; PT: public transits; EM: e-motorcycle.

Over the past two decades, the field of machine learning has seen significant advancements, leading to the emergence of several new techniques that enhance model performance. One noteworthy development in the data science and machine learning community was the Netflix Prize, held from 2006 to 2009 to predict Netflix user ratings on various movies (Bennett & Lanning, 2007). The winning team and other top-performing teams gained valuable insights into developing high-performance machine learning models. According to descriptions provided by the two top teams (Koren, 2009; Töscher et al., 2009), these breakthroughs were mainly attributed to three significant aspects: discovering new features underlying the data; employing different types of base learners; and blending predictions from different algorithms through ensemble methods such as gradient-boosted DT. Following the Netflix Prize, numerous competitions have been conducted over the past decade on platforms such as Kaggle (Bojer & Meldgaard, 2021) for various data sets and tasks. One popular category of ensemble methods is to blend different algorithms together, known as the stacking method (Hastie et al., 2009a), which is frequently employed in diverse classification tasks such as detecting phishing websites (Li et al., 2019), identifying defective products on production lines (Mangal & Kumar, 2016), and predicting antigenic variants of H1N1 influenza (Yin et al., 2018).

Given the constraints of prior freight choice models and the recent advances in developing high-performance models, this study aims to construct a high-performance model using solely the CFS PUF data set without any external data sources. The objective is to build a model framework that achieves superior performance by incorporating the three aspects gleaned from the Netflix Prize teams and other recent machine learning modeling practices into the development process.

In this study, we considered two distinct ensemble learning techniques: the weighted voting method and the stacking method. For feature extraction, we identified routed distances (estimated given the combination of origin, destination, and mode) as a crucial factor. Additionally, other features such as value-to-weight ratio and origin and destination metropolitan type (i.e., combined statistical areas, metropolitan statistical areas, and other regions) were also found useful. To construct base learners of the stacking and voting methods, we developed local models for various Standard Classification of Transported Goods types (SCTG code) and North American Industry Classification System types (NAICS code). Finally, stacking and voting methods were applied to ensemble the base learners from different commodity and industry genres and data scopes. Given these countermeasures, our findings indicate that all three countermeasures were effective in improving model performance. To the best of the authors knowledge, this is the first machine learning model that achieved accuracies above 90% using the CFS Public Use Microdata data set on freight mode choice predictions. This powerful model not only generates the most probable mode of transport but also estimates the probability of selecting each mode. In summary, this predictive model driven by machine learning can assist various stakeholders such as courier agencies, shippers, and transportation agencies in analyzing freight-related issues across different geographic scales (e.g., state-level, national-level) and product types (e.g., commodity type, industry type).

# 3 Explorative Data Analysis

## 3.1 Mode Classification and Variable Definitions

To compare the model performance of this study to the results of the previous work, this study used the five categories of freight modes that were adopted by two previous studies (Keya et al., 2021; Uddin et al., 2021). **Table 2** describes the five modes used in this study, along with the original 13 types of major freight modes defined by the CFS. In this classification, rail, water, and pipeline modes are combined into a single mode referred to as other modes. Thus, the 13 modes from CFS are compressed to 5 freight modes: for-hire truck, privately owned truck, parcel service, air, and other modes. Privately owned trucks include company-owned trucks used for logistics activities, and for-hire trucks include self-employed drivers transporting goods over long distances. In total, 18,539 (0.31%) out of 5,978,523 records are provided at a more aggregated level (e.g., single mode) and are not matched to a specific mode in **Table 2**. Therefore, this study excluded such records and analyzed 5,959,984 (99.69%) records of the 2017 CFS PUF data.

**Table 2. Five aggregated freight mode classifications and more specific CFS PUF data set mode classifications**

| Mode classification | Code defined by the CFS PUF data set |
|---|---|
| **For-hire truck (Mode 1)** | |
| For-hire truck | 04 |
| **Privately owned truck (Mode 2)** | |
| Privately owned truck | 05 |
| **Parcel service (Mode 3)** | |
| Parcel service (parcel, USPS, or courier) | 14 |
| **Air (Mode 4)** | |
| Air (including truck and air) | 11 |
| **Other modes (Mode 5)** | |
| Rail | 06 |
| Water | 07 |
| Inland water (water) | 08 |
| Great lakes (water) | 09 |
| Deep sea (water) | 10 |
| Multiple waterways (water) | 101 |
| Pipeline | 12 |
| Truck and rail (multiple modes) | 15 |
| Truck and water (multiple modes) | 16 |
| Rail and water (multiple modes) | 17 |

**Table 3** presents all information considered from the CPS PUF data set. There are three categories of information: basic shipment information, spatial information, and supplementary information. The basic shipment information includes weight (in US pounds), value (in US dollars), commodity type (SCTG), and industry type (NAICS). The spatial information is based on origin, destination, great circle distance, and the routed distance for each specific freight mode. The 2017 CFS PUF data set reports the routed distance estimated for only the chosen mode. Because of this dependency on the chosen mode, directly incorporating the routed distance as model inputs results in a loss of information, which may cause modeling bias and overfitting. To address this, this study utilized derived routed distance estimates by origin and destination for each of the five modes, including both chosen and unchosen modes (as detailed in **Section 4**). This approach allowed for constructing a geographical profile of various freight modes between states or metropolitan regions. The group of variables of all derived distances for all available modes were employed as model inputs. Supplemental information pertains to detailed data regarding shipments of specialized commodities that require specific handling conditions because of their hazardous nature (e.g., chemical compounds) or temperature sensitivity (e.g., fresh meat). Such information proves valuable in forecasting the appropriate shipping mode for such commodities. Notably, our machine learning model does not rely on external data sets, thereby easing the efforts of data collection for researchers in their analyses. For each piece of information, there exists one or multiple variables to describe it. In the context of machine learning, these input variables are referred to as input features.

**Table 3. Description of all considered information from the CFS PUF data set**

| Column name | Description |
| --- | --- |
| **Basic shipment information** | |
| Mode (target variable) | Shipment mode (for-hire truck, privately owned truck, parcel, air, other modes) |
| Weight | Weight of the shipment (in US pounds) |
| Value | Value of the shipment (in US dollars) |
| SCTG | Two-digit SCTG commodity code of the shipment (43 categories) |
| NAICS | Industry classification of the shipper (45 categories) |
| **Spatial information** | |
| Origin | Origin code (state code and metropolitan area code) |
| Destination | Destination code (state code and metropolitan area code) |
| Distance | Great circle distance between origin and destination (in miles) |
| Routed distance | For the chosen mode, the actual routed traveling distance of the shipment (in miles) |
| **Supplemental information** | |
| HAZMAT | Hazardous material (Class 3.0 hazard, other hazard, or not hazard) |
| Temperature controlled | Temperature-controlled shipment |

### 3.2 Analyzing Freight Mode Choice Patterns

**Figure 1** shows the distribution of the five freight modes across multiple metrics for the 2012 and 2017 CFS data sets. The four plots, from left to right, correspond to metrics based on shipment count, total weight (in US tons), total value (in US dollars), and total ton-miles traveled (in US ton-mile). Across metrics, the share of different modes varies significantly. For instance, although most shipments correspond to the parcel service parcel in terms of shipment numbers, parcel service only represents a very small share in terms of the total weight. Besides, a side-by-side comparison between the 2012 and 2017 data sets provides insight into trends observed over 5 years. Between these two years, the shares of other modes remained relatively stable except for an increase in parcel service and a decrease in privately owned trucks. For analysis in this study, only the 2017 data set was used.

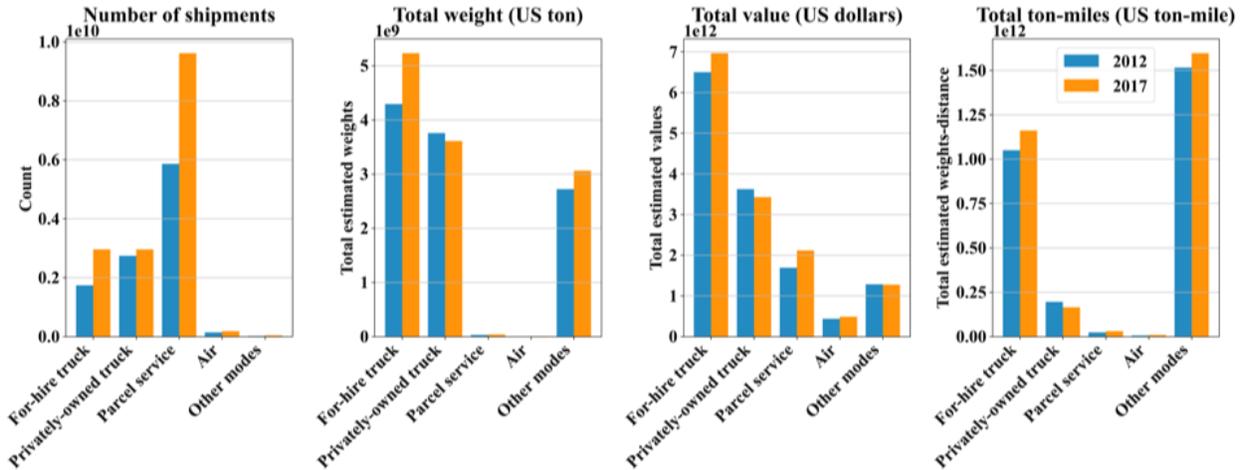

**Figure 1. Share of different modes with respect to the number of shipments, total weight, total value, and total ton-miles traveled**

Through the use of total values in US dollars as a metric, **Figure 2** and **Figure 3** illustrate the percentage distribution (in terms of value) of various modes of shipment origins and destinations in different US states. The results demonstrate that different states exhibit varying mode distributions. For instance, privately owned trucks are more prevalent in the Northeast, whereas for-hire trucks dominate many Midwest states. Furthermore, this study also used spatial information inferred from the derived distance between modes with respect to origin and destination states and metropolitan regions (**Section 4**).

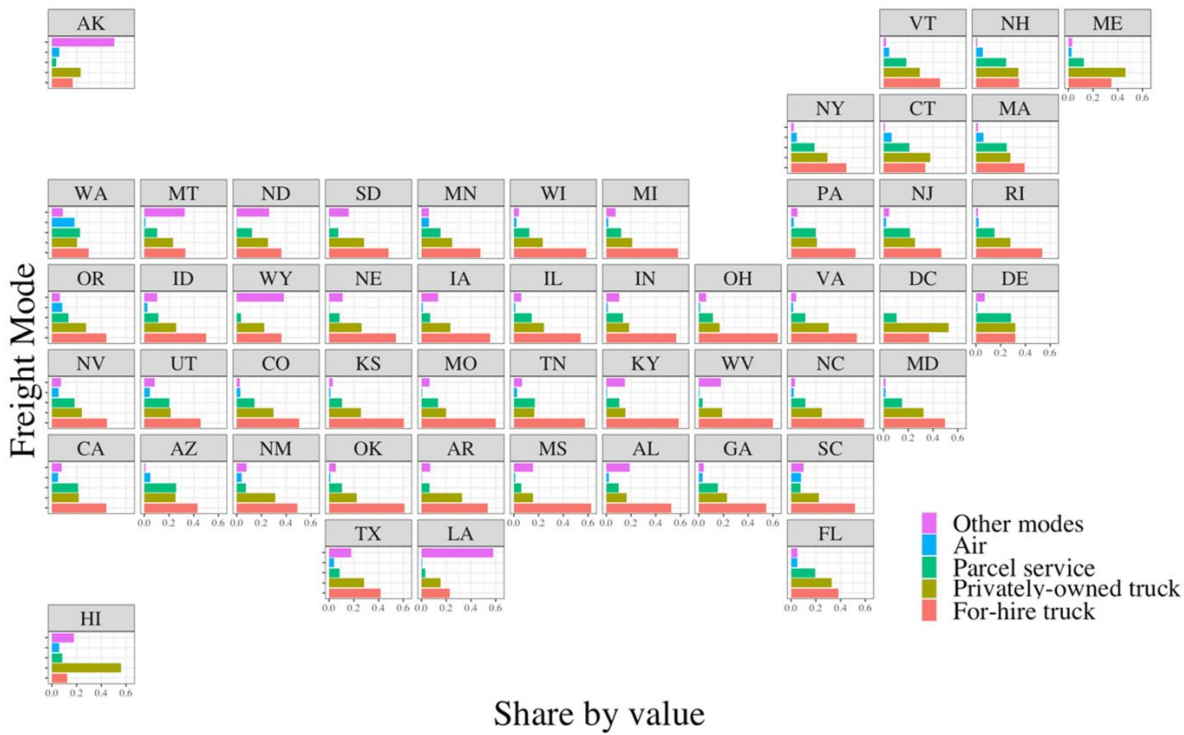

**Figure 2. Percentage value share (in US dollars) of different freight modes by origin state**

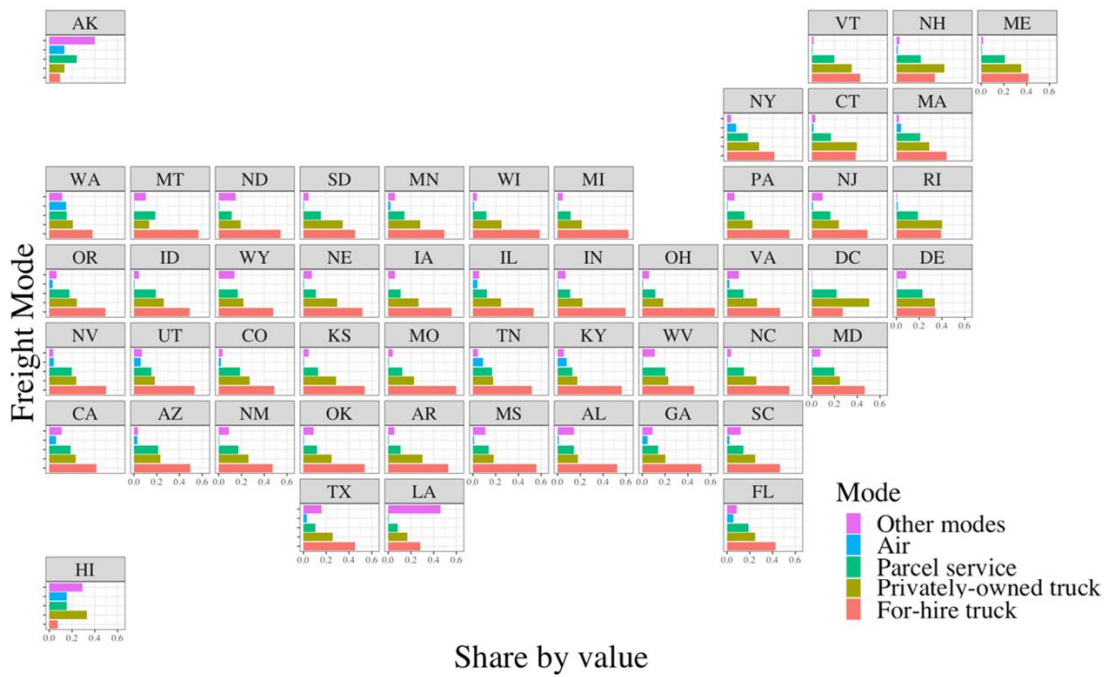

**Figure 3. Percentage value share (in US dollars) of different freight modes by destination state**

In addition to the spatial distribution of freight modes, the modal share of different product types must also be understood. To this end, we present the results of our distribution analysis in **Figure 4**, which displays the modal split by each commodity type according to the SCTG code and by each industry type according to the NAICS code. It is found that certain commodities present distinct mode preferences. For instance, alcoholic beverages (SCTG #08) are predominantly transported by privately owned trucks because of their requirement for special permits (Alcohol Transport Permit Requirements by State | Brew Movers, 2022). Meanwhile, parcel services are commonly used for transporting electronic devices (SCTG #35) or precision instruments (SCTG #38), likely because of their small size/weight and urgent delivery requirements.

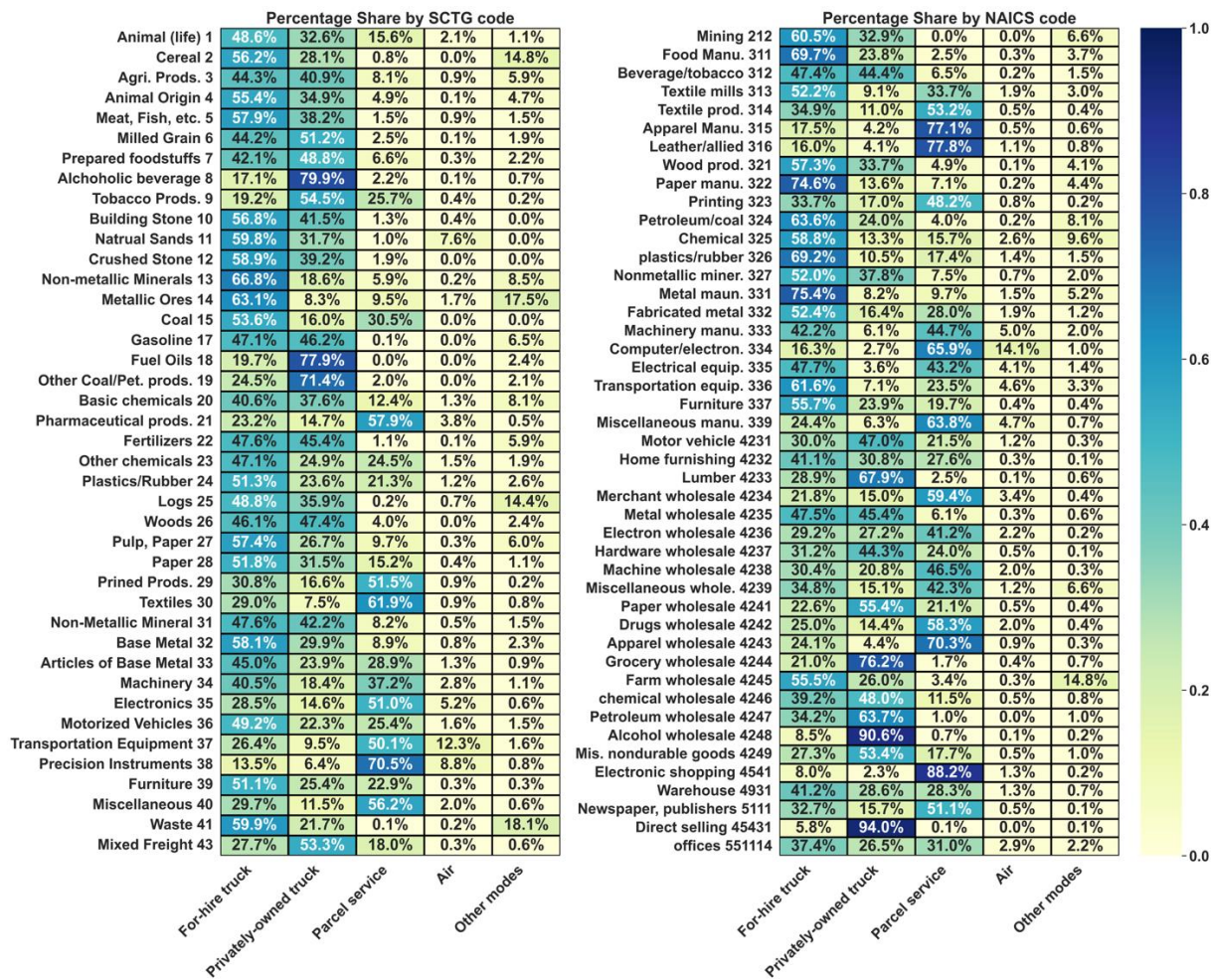

**Figure 4. Distribution of different freight mode choices by different SCTG (commodity) codes and NAICS (industry) codes (measured in the number of shipments)**

In summary, this section presents the results of an exploratory data analysis aimed at identifying key factors and trends to guide the modeling step in freight mode selection. First, the analysis reveals that different locations exhibit varying proportions of shipments. Therefore, incorporating both geographical traits and travel distances between origin and destination is crucial in narrowing the feasible freight modes. Second, the SCTG and NAICS codes characterizing commodity types play a significant role in selecting the preferred freight mode. Given the existence of more than 40 categories for SCTG and NAICS types, careful consideration must be given to exploit such categorical heterogeneity across different cases. Finally, the distribution of different freight modes varies significantly under different metrics. Consistent with previous studies (Keya et al., 2018; Rich et al., 2009; Uddin et al., 2021), we employed the number of shipments (i.e., records) for model evaluation.

## 4  Methodology

The structure of the methodology subsections is delineated as follows: (1) **Section 4.1** outlines the general data processing pipeline employed for constructing the model; (2) **Section 4.2** develops the augmented data features used for machine learning models; (3) **Section 4.3** shows different types of level-1 model as well as the benefits of developing local/separate models to exploit the heterogeneous nature of different commodities/industries types of goods; (4) **Section 4.4** shows the level-2 model blending different genres of level-1 model via model stacking/voting techniques; Furthermore, more detailed steps for training level-2 stacking methods are provided in this section.

## 4.1 An Overview of the Proposed Methodology

To develop a high-performance predictive model, it is essential to construct an efficient data pipeline through an analysis framework. This study focuses on building pipelines to prevent overfitting, data leakage issue, and to compare against different machine learning models and layouts. **Figure 5** illustrates the data processing pipeline and analysis framework employed in this study. Step 1 in the pipeline is to split the data into a training set and a testing set. The testing set comprises 20% of the data, using stratified random sampling, grouped by SCTG code and NAICS code. In Step 2, useful geographical features are extracted, including great circle distance reported from the 2017 CFS PUF data and the derived distances between each origin and destination CFS metropolitan area (i.e., CFA area) for each of the five modes of transportation. At this point, the training and testing data sets are cleaned and formed. In Step 3, multiple machine learning models are trained and tested using different input features. Global base learners are trained using randomly drawn samples from all training data sets while separate models (base learners) for each commodity and industry type are trained and concatenated as one predicting model (i.e., base learner). In Step 4, different types of base learners are combined using either voting averages or stacking methods to create a final estimator (i.e., level-2 model or meta-learner). Each step of the data pipeline is discussed in further detail in the following sections.

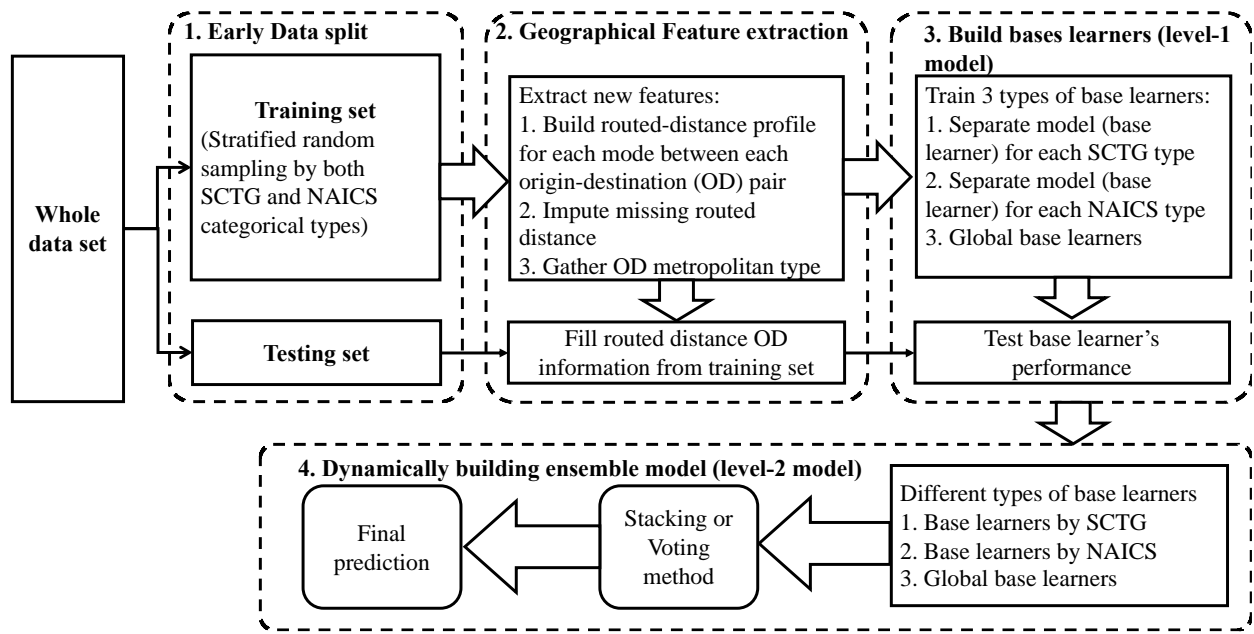

**Figure 5. Proposed data processing pipelines and analysis framework**

## 4.2 Developing Useful Features (Data Augmentation)

Routed distance is the actual distance traveled by a specific freight mode. To obtain useful geographical and distance information (i.e., step 2 of the framework), regional definitions must be well understood. The CFS PUF data set classifies the United States into more than 123 regions, including combined statistical areas, metropolitan statistical areas, and the remainder of each state (2017 Commodity Flow Survey (CFS) Public Use File (PUF) Data Users Guide, 2020). These regions divide the United States into distinct parts (i.e., CFS areas). Since the CFS records provide the estimated routed distances only for the selected mode, directly using the derived distance information in the model may lead to a loss of information. However, since the data set has millions of records, between a specific pair of origins and destinations, it is plausible to assume that there would be many records for each freight mode. Thus, all five modes' routed distances can be used jointly as input features in the model. For example, as input features, given a specific pair of origin and destination, these derived distances by each mode can be computed as a statistic (e.g., median) of observed routed distances.

In **Figure 6**, the map displayed on the left illustrates the identified CFS regions and their corresponding codes. Each yellow region represents a CFS area, and the remaining regions of each state are depicted in white and have a different code. The routed distance for various freight modes can vary significantly between an origin–destination pair. For instance, **Figure 6** shows that the routed distance by air between Los Angeles, California, and New York, New York, is shorter than that of for-hire trucks, which is still shorter than that of other modes (e.g., water, rail, pipes). Comparing the routed distances of various freight modes (top right of **Figure 6**) provides valuable information regarding mode choice. Notably, using routed/derived distance entails a potential risk of overfitting due to data leakage. In this case, the data leakages refer to the misbehavior of disclosing the predicting mode by providing the routed distance for the selected freight mode. Consequently, all routed distances must be generated as input features from the training data set. Moreover, in this study, only the median distances for each mode between origin–destination pairs are employed to describe a general derived distance given the candidate mode, origin, and destination. Also, it is likely that some modes are not available between some origins and destinations. For example, although privately owned trucks mode is theoretically possible between Los Angeles and New York, no such records are found in the dataset because the trip distance is too long (the privately owned trucks rarely ship beyond 500 miles). For such cases, we impute that derived distances of the missing modes by fitting a regression model to estimate the routed distance given the great circle distance.

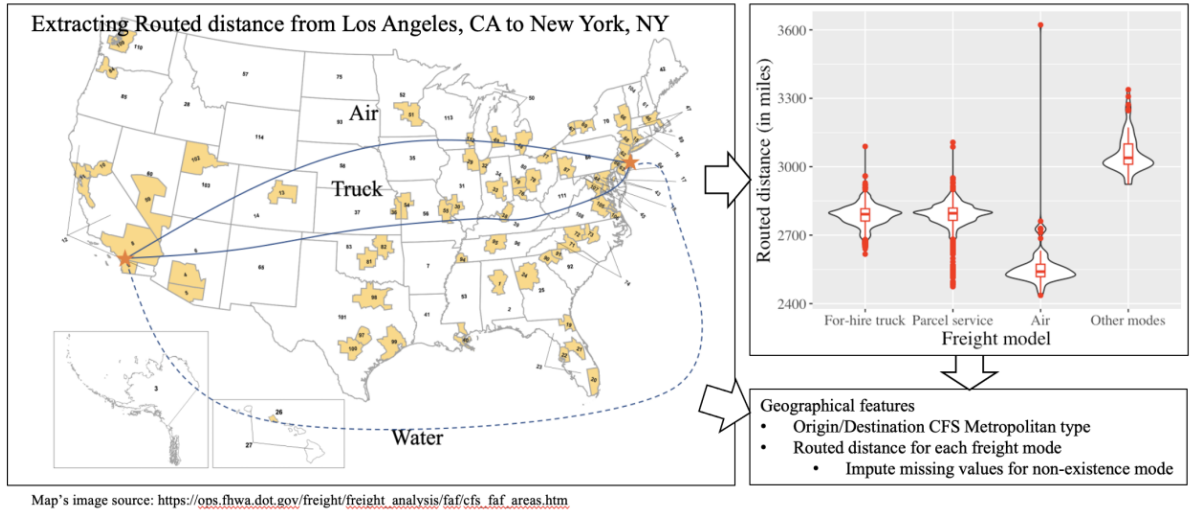

**Figure 6. Collected derived distance and spatial information for each mode between each origin–destination pair of CFS areas**

All explanatory input variables adopted in the model are described in **Table 4**, which were further derived from the data set inputs presented in **Table 3**. Compared with **Table 3**, the derived variables (marked by the asterisks in **Table 4**) are explained as follows. The value to weight variable, by measuring the products value given weights, is a useful derived feature. Origin and destination types are referred to by CFS area classes. Three classes are available: combined statistical area, metropolitan statistical area, and remaining part of each state. Although listed in **Table 3**, origin and destination CFS area is not directly fed into any predicting model. The origins and destinations are applied in an indirect way by aggregating derived distances between origin–destination pairs. Specifically, "M1"–"M5" correspond to the routed distance from each of the five modes. Missing values in routed distance for a specific mode (e.g., "M5") between an origin–destination pair can arise from limited coverage of the data records, and it is imperative to impute these values for model execution. To this end, a linear regression model was constructed between the great circle distance and derived distance to impute those missing derived distances. Since imputation of missing values may affect model performance, we also propose a set of features to describe whether each derived distance has been imputed or not using Boolean values (true or false), denoted as "I1"–"I5" for five specific modes.

**Table 4. All explanatory input variables and features employed in the model**

| Variable | Feature | Description |
| --- | --- | --- |
| **Basic shipment information** | | |
| Single weight | "sw" | Weight of the shipment (in US pounds) |
| Single Value | "sv" | Value of the shipment (in US dollars) |
| SCTG | "sctg" | Two-digit SCTG commodity code of the shipment (43 categories) |
| "sctg_n" | "sctg_n" | Nine types of aggregated SCTG commodity codes |
| NAICS | "naics" | Industry classification of the shipper (45 categories) |
| *Value to weight ratio | "v2w" | Value to weight ratio for the shipment (US dollars per pound) |
| **Spatial information** | | |
| Origin | "orig" | Origin CFS area code |

| *Origin area type | "orig_type" | Three classes of CFS areas (C: combined statistical area; M: metropolitan statistical area; R: remaining part of each state) |
|---|---|---|
| Destination | "dest" | Destination CFS area code |
| *Destination area type | "dest_type" | Three classes of CFS areas (C: combined statistical area; M: metropolitan statistical area; R: remaining part of each state) |
| Great Circle Distance | "dc_dist" | Great circle distance between origin and destination (in miles) |
| *Routed Distance | "M1," "M2," "M3," "M4," "M5" | For each freight mode, the median routed traveling distance between origin and destination (in miles) |
| *Imputed Distance | "I1," "I2," "I3," "I4," "I5" | For each freight mode, whether the routed traveling distance was imputed since no records are recorded between origin and destination (in miles) |
| **Supplemental information** | | |
| HAZMAT | "haz" | Hazardous material (Class 3.0 hazard, other hazard, not hazard) |
| Temperature controlled | "temp_cntl" | Temperature-controlled shipment |

*Derived information

## 4.3 Building Base-learner (Level-1 Model)

In addition to expanding the dataset by deriving more features from existing information, performance enhancement can also be achieved by constructing distinct sub-models tailored to specific data categories. Observed from **Figure 4**, different commodity or industry types exhibit different preferences and patterns in freight modes. Moreover, dedicated models are frequently fitted to model a specific type of commodity (e.g., hazardous commodity) in some previous studies (Nowlan et al., 2021; Xie et al., 2012). Those practical instances imply that pursuing a single for all product categories may not lead to an optimal performance. Given the limited number of 43 SCTG types and 45 NAICS types, it is entirely feasible to build a dedicated model by each SCTG and each NAICS category. Even for the category with fewest records, there are still enough number of records (e.g., over 2,000) to fit base learners.

In conclusion, three distinct modeling approaches can be considered: (1) a unified base learner applied on all record types; (2) a dedicated model for each SCTG type; (3) a dedicated model for each NAICS type. Fitting dedicated models for different types are not equivalent to the act of adding NAICS/SCTG features (e.g., one-hot encoding variables for SCTG type or NAICS type) for a unified model. Our base learners are trained utilizing a diverse set of machine learning algorithms, encompassing LR, DT, SVM, MLP, KNN, NB, RF, Gradient Boosted Decision Tree (GB), AdaBoost (ADA), and Extra Trees (Extra). Importantly, for any machine learning models, all three types of models can be built as base learners. For any trip to predict, one can dynamically get results from three types of models given its SCTG/NAICS type. Although any available machine learning models can be used as base-learners, only those best performed models are suitable as candidates for level-2 model, as elaborated in the following section.

## 4.4 Dynamically Building Ensemble Models (Level-2 model)

Voting mechanisms are commonly employed ensemble techniques for generating a final prediction by aggregating the outputs of multiple models (Hastie et al., 2009b). Moreover, a diverse range of local models must be considered that specialize in modeling different commodities and industry types. To consider such models, we propose a dynamic voting scheme, as illustrated in **Figure 7**. Specifically, all relevant base learners/models are selected for each record based on their commodity and industry types. These local models, together with a global base learner, constitute a pool of models. Each model generates five probabilities for selecting five freight modes. A voting classifier then computes the average probability of selecting each mode across all models. The predicted freight mode with the highest probability is considered the final outcome.

---
**Algorithm 1.** Dynamically combining different base learners using a voting mechanism

1:   **procedure** Collect_Models_For_One_Record(record, model_types)
2:     $S_{M1}$ = Get_Base_Learners_By_SCTG(record.sctg_id, model_types);
3:     $S_{M2}$ = Get_Base_Learners_By_NAICS(record.naics_id, model_types);
4:     $S_{M3}$ = Get_Global_Learners(model_types);
5:     $S_{pool} = \{s: s \in S_{M1} \cup S_{M2} \cup S_{M3}\}$;
6:     $V_{pool}$ = Compute_Predict_Vectors_For_Each_Learner($S_{pool}$);
7:     $V_{final}$ = Get_Average_Voting_By_Probabilities($V_{pool}$);

---

**Figure 7. One possible pseudocode of steps applying a voting mechanism to obtain ensembled predictions of base learners for each record.**

**Figure 8** illustrates the stacking methodology employed in this study. Analogous to the voting approach, high-performing tree-based models are employed as level-1 base learners. To obtain predictions for the level-2 model, an XGBoost model is trained on the probability predictions and a subset of input features from the level-1 models (Chen & Guestrin, 2016). Compared to other models like RF, XGBoost model has the benefits of fitting on large scale data more efficiently. In contrast, XGBoost model is not used for level-1 model in this study because the model takes too much memory space compared to other machine learning models. The purpose of the level-2 model is to determine optimal weights for the different combinations of NAICS and SCTG types, utilizing a DT model to achieve this objective. Finally, five probabilities are generated as predictions corresponds to each of the five different freight modes. Although the final mode prediction is determined by selecting the prediction with the highest probability, these five probabilities can also provide useful information since products are often distributed using more than one freight modes.

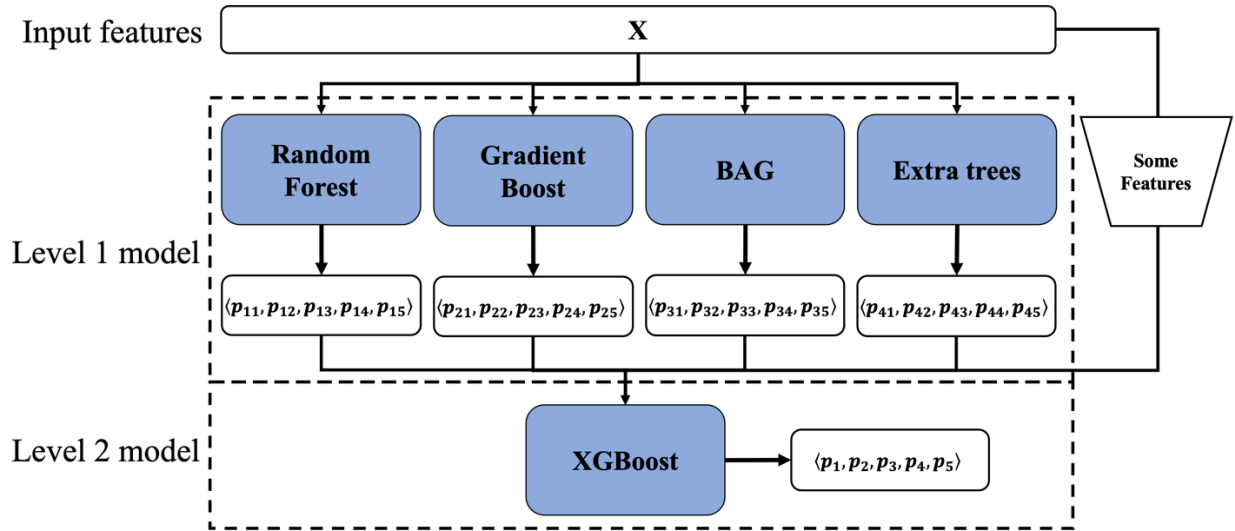

**Figure 8. The basic structure of stacking different types of tree-based methods using a two-level hierarchy to obtain final ensembled predictions**

The precise steps to generate the training and testing data sets considering 2-level stacking methods are summarized in **Figure 9.** As previously mentioned, the first data preparation step is to do a stratified sampling approach to reserve 20% of the data as the testing data set. To prevent data leakage of derived features, the derived features for the testing set were only derived from the training set instead of testing set (**Figure 9** (bottom left)). Furthermore, to avoid data leakage during the data preparation step before training, a one out-of-bag method was used to generate derived features (**Figure 9** (bottom right)). This method ensures that no data records used routed distances from their own group to obtain derived distance.

Notably, derived features only needed to be computed once during the entire data pipeline. When using the stacking method, a level-2 meta-learner is required (**Figure 9** (top right)). The inputs for the meta-learner were generated from the validation set of each *k*-fold model to prevent overfitting (orange blocks in **Figure 9**). During level-1 model training, a five-fold cross-validation approach was applied to determine the best-performing base learners with respect to their meta-parameters. For level-2 models, all training data sets were used to train a meta-learner and were then tested using the reserved testing data set. This means that final model delivered was tested on the testing data set.

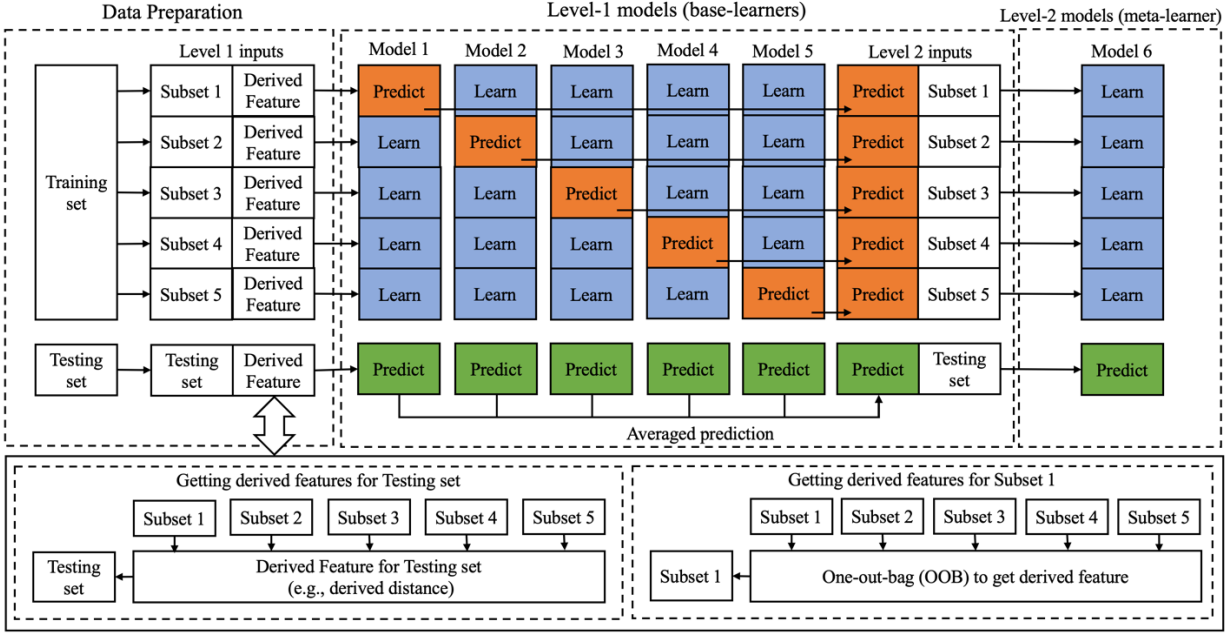

**Figure 9. Data flow of the analysis framework: data preparation and cross-validation for the two-level stacking method**

## 5   Experiment Results

### 5.1   Model Implementation

To assess the efficacy of various methodologies, an extensive range of machine learning models were employed in this study. Similar to some of the previous studies in **Table 1**, the models used encompassed Logistics Regression (LR), DT, SVM, MLP, KNN, NB, RF, GB, ADA, and extra trees (Extra). In most cases, the Scikit-learn package in Python was used to execute the models (Pedregosa et al., 2011). For the stacking model, the XGBoost package in Python was preferred for analysis because of its greater computational efficiency compared with the Scikit-learn package (Chen & Guestrin, 2016).

### 5.2   Experiment Design and Model Evaluation

In addition to the model performance, the input features and data fitting scope of the model were also evaluated. To enhance the model efficacy, this study proposes three countermeasures: (1) use derived distance as input features; (2) build dedicated models for each SCTG/NAICS type; (3) use ensemble methods to combine different models. The experimental controls for each proposed countermeasure are detailed in **Table 5**. More specifically, to compare the effectiveness of each of the three countermeasures, seven scenarios were developed and are presented in **Table 6**. The training set for all data sets comprises 80% of the total samples, and the remaining 20% of samples are reserved as testing set. Testing samples were selected randomly within each NAICS and SCTG category.

**Table 5. Experiment design aspects to measure the effectiveness of three countermeasures**

| Input features | Description |
| --- | --- |
| "without derived distance" | "M1"–"M5" and "I1"–"I5" are not considered in the model |
| "with derived distance" | "M1"–"M5" and "I1"–"I5" are considered in the model |
| **Modeling scope** | |
| "unified model" | Only one model is built across the entire data set |
| "local (concatenated) model" | One separate model is built for each NAICS and SCTG category |
| **Stacking method** | |
| "voting model" | The average of output probabilities of other models is used as the final prediction |
| "stacking model" | An ensemble structure is applied (**Figure 8**) based on other models |

To effectively compare different models, various evaluation metrics were computed. These metrics include accuracy, balanced accuracy (BA), precision, recall, and F1 score. Simply measuring accuracy alone may not accurately represent the distribution of data between the air and other modes because they constitute only a small portion of the overall data set (as illustrated in **Figure 10**). To address this issue, the BA considered the accuracy by averaging them across all five modes to mitigate the impacts of data imbalance issue. All metrics were measured based on the number of records without any further weighting (e.g., by values) to ensure comparability with previous studies.

Based on the analysis, the stacking model was selected as the final model. To further examine the stacking model performance, **Figure 10** displays the results by SCTG code (left) and NAICS code (right). Overall, the final model proved effective across most SCTG and NAICS types, except for the air mode for some SCTG and NAICS types because of a severe data imbalance.

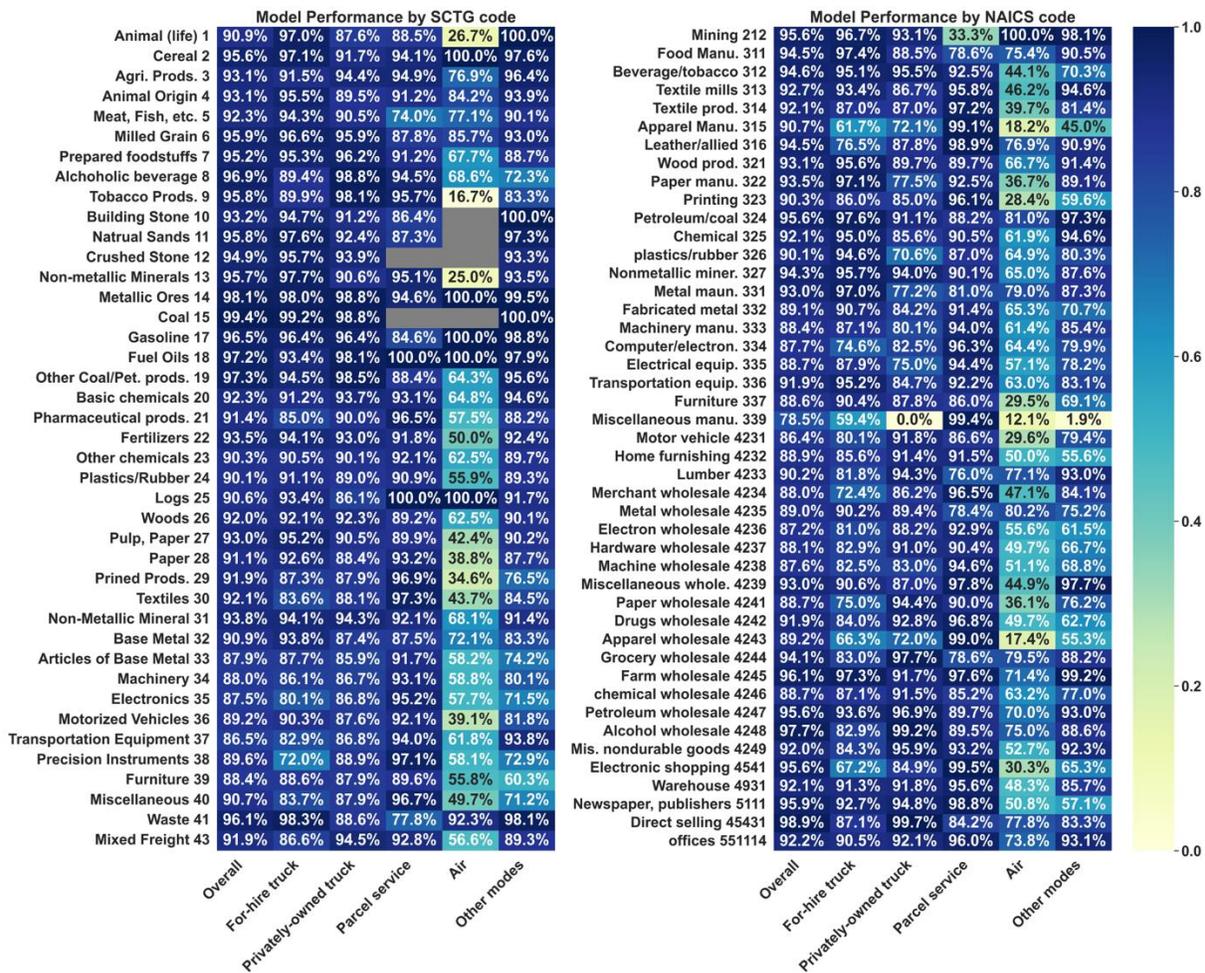

**Figure 10. Accuracy of the final model for each freight mode by SCTG code (the left column) and by NAICS code (the right column)**

Furthermore, by adjusting the probability threshold for each mode in decision-making, a receiver operating characteristic curve was generated with the corresponding Area Under the Curve (AUC) score reported in **Figure 11**. For the AUC score, the closer to 1, the better the model performed. By comparison, ensemble methods like stacking or voting model (the third row of subplots) can improve the AUC score by combining results from different base learners (the first two rows of subplots). **Figure 11** will be discussed in more detailed way in the following section.

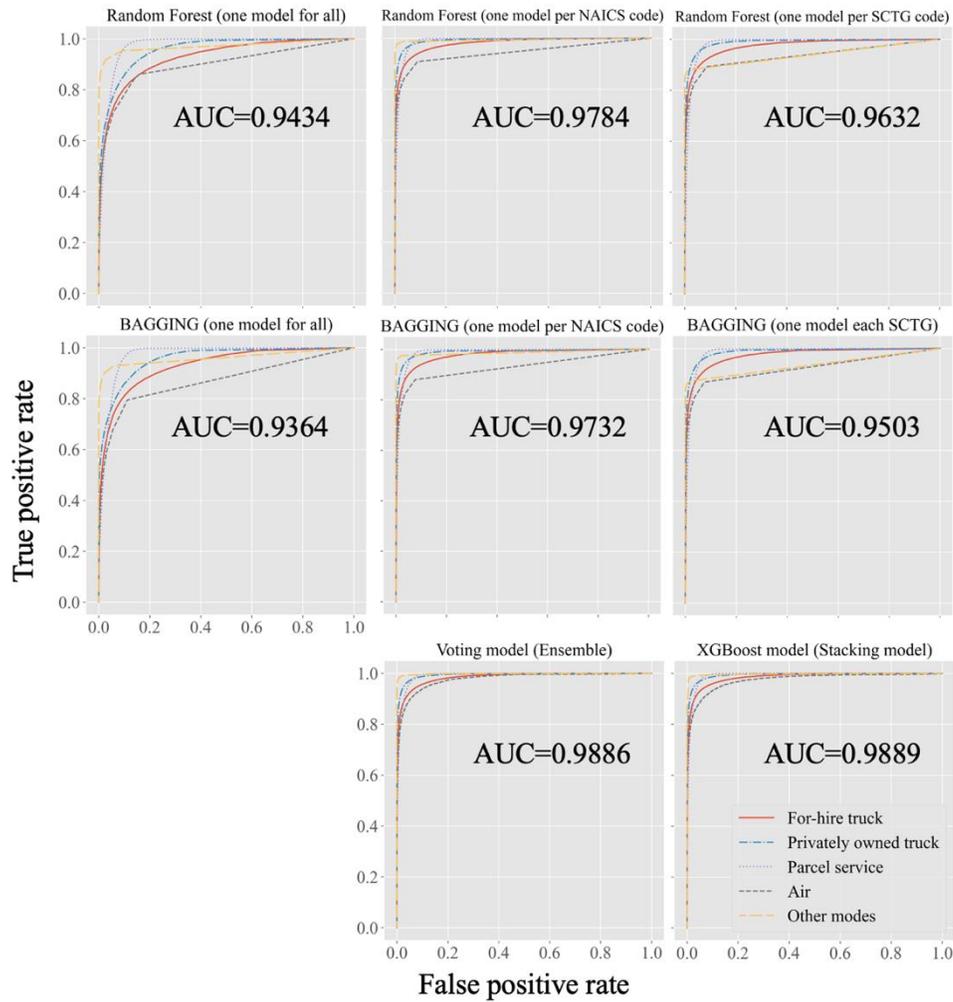

**Figure 11. Evaluating performance by receiver operating characteristic's AUC for various models**

### 5.3 Model Results

**Table 6** presents the results of evaluating various models under different specifications. Our findings demonstrate that the performance of these models can be significantly enhanced by implementing three key improvements: incorporating the distance between origin and destination for each freight mode; developing local models for each category; and adopting a voting/ensemble approach to combine different types of local and unified models.

**Table 6. Performance of unified models, SCTG models, NAICS models, and the overall voting model on the testing data set (boldface numbers are the best of each row)**

| | LR | DT | SVM | MLP | KNN | NB | RF | GB | BAG | ADA | Extra | XGBoost | Vote | Stack |
|---|---|---|---|---|---|---|---|---|---|---|---|---|---|---|
| (a) Unified models (without derived distance, trained on 10,000 training records) | | | | | | | | | | | | | | |
| Accuracy | 0.559 | 0.653 | 0.168 | 0.709 | 0.630 | 0.439 | 0.729 | 0.670 | 0.710 | 0.610 | 0.668 | 0.713 | **0.731** | 0.726 |
| BA | 0.395 | 0.579 | 0.283 | 0.583 | 0.490 | 0.238 | 0.623 | 0.591 | 0.612 | 0.508 | 0.444 | 0.600 | 0.617 | **0.627** |
| Precision | 0.553 | 0.653 | 0.437 | 0.717 | 0.628 | 0.468 | 0.731 | 0.676 | 0.708 | 0.622 | 0.686 | 0.713 | **0.733** | 0.726 |
| Recall | 0.559 | 0.653 | 0.168 | 0.709 | 0.630 | 0.439 | 0.729 | 0.670 | 0.710 | 0.610 | 0.668 | 0.713 | **0.731** | 0.726 |
| F1 | 0.541 | 0.653 | 0.181 | 0.701 | 0.627 | 0.356 | 0.725 | 0.664 | 0.708 | 0.598 | 0.659 | 0.709 | **0.728** | 0.720 |
| $\hat{\sigma}$(Accuracy) | 0.000 | 0.070 | 0.046 | 0.052 | 0.047 | 0.060 | 0.000 | 0.004 | 0.009 | 0.027 | 0.026 | 0.001 | 0.000 | 0.000 |
| (b) Unified models (with derived distance, trained on 10,000 training records) | | | | | | | | | | | | | | |
| Accuracy | 0.563 | 0.779 | 0.374 | 0.787 | 0.681 | 0.511 | 0.819 | 0.716 | 0.824 | 0.595 | 0.796 | 0.821 | 0.837 | **0.841** |
| BA | 0.469 | 0.696 | 0.210 | 0.709 | 0.555 | 0.333 | 0.699 | 0.561 | 0.709 | 0.491 | 0.671 | 0.724 | 0.715 | **0.752** |
| Precision | 0.560 | 0.779 | 0.366 | 0.789 | 0.682 | 0.510 | 0.818 | 0.724 | 0.822 | 0.615 | 0.797 | 0.821 | 0.837 | **0.841** |
| Recall | 0.563 | 0.779 | 0.374 | 0.787 | 0.681 | 0.511 | 0.819 | 0.716 | 0.824 | 0.595 | 0.796 | 0.822 | 0.837 | **0.840** |
| F1 | 0.547 | 0.779 | 0.303 | 0.784 | 0.680 | 0.499 | 0.817 | 0.708 | 0.821 | 0.580 | 0.794 | 0.820 | 0.834 | **0.839** |
| $\hat{\sigma}$(Accuracy) | 0.000 | 0.095 | 0.074 | 0.089 | 0.078 | 0.074 | 0.000 | 0.058 | 0.055 | 0.103 | 0.095 | 0.000 | 0.000 | 0.000 |
| (c) Concatenate local models for each NAICS code (without derived distance, trained on all training data) | | | | | | | | | | | | | | |
| Accuracy | 0.676 | 0.752 | 0.351 | 0.759 | 0.774 | 0.602 | 0.797 | 0.759 | 0.797 | 0.715 | 0.756 | 0.786 | **0.806** | 0.785 |
| BA | 0.499 | 0.715 | 0.345 | 0.649 | 0.716 | 0.398 | 0.723 | 0.650 | 0.739 | 0.529 | 0.624 | 0.681 | **0.735** | 0.722 |
| Precision | 0.671 | 0.752 | 0.507 | 0.759 | 0.773 | 0.611 | 0.796 | 0.758 | 0.796 | 0.711 | 0.756 | 0.787 | **0.806** | 0.786 |
| Recall | 0.676 | 0.752 | 0.351 | 0.759 | 0.774 | 0.602 | 0.797 | 0.759 | 0.797 | 0.715 | 0.756 | 0.786 | **0.806** | 0.785 |
| F1 | 0.670 | 0.752 | 0.387 | 0.755 | 0.773 | 0.594 | 0.796 | 0.756 | 0.796 | 0.709 | 0.753 | 0.783 | **0.805** | 0.783 |
| $\hat{\sigma}$(Accuracy) | 0.001 | 0.070 | 0.164 | 0.001 | 0.008 | 0.077 | 0.000 | 0.019 | 0.018 | 0.034 | 0.030 | 0.000 | 0.000 | 0.000 |
| (d) Concatenate local models for each NAICS code (with derived distance, trained on all training data) | | | | | | | | | | | | | | |
| Accuracy | 0.681 | 0.884 | 0.441 | 0.843 | 0.873 | 0.636 | 0.916 | 0.812 | 0.913 | 0.712 | 0.902 | 0.868 | **0.922** | 0.919 |
| BA | 0.552 | 0.834 | 0.325 | 0.752 | 0.818 | 0.482 | 0.849 | 0.696 | 0.848 | 0.530 | 0.837 | 0.775 | **0.859** | **0.859** |
| Precision | 0.679 | 0.884 | 0.476 | 0.842 | 0.873 | 0.634 | 0.916 | 0.812 | 0.913 | 0.710 | 0.903 | 0.870 | **0.923** | 0.920 |
| Recall | 0.681 | 0.884 | 0.441 | 0.843 | 0.873 | 0.636 | 0.916 | 0.812 | 0.913 | 0.712 | 0.902 | 0.868 | **0.922** | 0.919 |
| F1 | 0.678 | 0.884 | 0.447 | 0.841 | 0.873 | 0.632 | 0.915 | 0.809 | 0.913 | 0.706 | 0.902 | 0.866 | **0.922** | 0.918 |
| $\hat{\sigma}$(Accuracy) | 0.000 | 0.101 | 0.073 | 0.000 | 0.015 | 0.105 | 0.000 | 0.052 | 0.048 | 0.084 | 0.080 | 0.001 | 0.000 | 0.000 |
| (e) Concatenate local models for each SCTG code (without derived distance, trained on all training data) | | | | | | | | | | | | | | |
| Accuracy | 0.661 | 0.724 | 0.280 | 0.742 | 0.755 | 0.554 | **0.773** | 0.750 | 0.769 | 0.706 | 0.724 | 0.766 | 0.759 | 0.751 |
| BA | 0.505 | 0.689 | 0.294 | 0.639 | 0.685 | 0.363 | **0.700** | 0.641 | 0.713 | 0.545 | 0.582 | 0.659 | 0.488 | 0.486 |
| Precision | 0.658 | 0.724 | 0.415 | 0.742 | 0.755 | 0.572 | **0.773** | 0.750 | 0.768 | 0.704 | 0.729 | 0.768 | 0.751 | 0.746 |
| Recall | 0.661 | 0.724 | 0.280 | 0.742 | 0.755 | 0.554 | **0.773** | 0.750 | 0.769 | 0.706 | 0.724 | 0.766 | 0.759 | 0.751 |
| F1 | 0.655 | 0.724 | 0.306 | 0.739 | 0.754 | 0.545 | **0.771** | 0.746 | 0.768 | 0.701 | 0.72 | 0.762 | 0.754 | 0.744 |
| $\hat{\sigma}$(Accuracy) | 0.000 | 0.081 | 0.190 | 0.000 | 0.006 | 0.092 | 0.000 | 0.012 | 0.010 | 0.027 | 0.026 | 0.001 | 0.000 | 0.000 |
| (f) Concatenate local models for each SCTG code (with derived distance, trained on all training data) | | | | | | | | | | | | | | |
| Accuracy | 0.662 | 0.869 | 0.423 | 0.867 | 0.867 | 0.586 | 0.902 | 0.806 | 0.901 | 0.706 | 0.887 | 0.853 | **0.908** | **0.908** |
| BA | 0.563 | 0.820 | 0.340 | 0.814 | 0.814 | 0.430 | 0.836 | 0.691 | **0.838** | 0.546 | 0.819 | 0.759 | 0.834 | 0.833 |
| Precision | 0.662 | 0.869 | 0.467 | 0.866 | 0.866 | 0.585 | 0.902 | 0.807 | 0.901 | 0.707 | 0.888 | 0.855 | 0.908 | **0.910** |
| Recall | 0.662 | 0.869 | 0.423 | 0.867 | 0.867 | 0.586 | 0.902 | 0.806 | 0.901 | 0.706 | 0.887 | 0.853 | 0.908 | **0.909** |
| F1 | 0.659 | 0.869 | 0.423 | 0.866 | 0.866 | 0.581 | 0.901 | 0.803 | 0.900 | 0.701 | 0.886 | 0.851 | 0.907 | **0.908** |
| $\hat{\sigma}$(Accuracy) | 0.001 | 0.119 | 0.126 | 0.001 | 0.010 | 0.128 | 0.000 | 0.048 | 0.045 | 0.081 | 0.076 | 0.001 | 0.000 | 0.000 |

(g) Ensemble model combining (b)–(d) (with and without derived distance, trained on all training data)

| | Vote | | Stack | |
|---|---|---|---|---|
| | Without derived distance | With derived distance | Without derived distance | With derived distance |
| Accuracy | 0.851 | 0.923 | 0.862 | **0.928** |
| BA | 0.755 | 0.851 | 0.798 | **0.871** |
| Precision | 0.854 | 0.924 | 0.861 | **0.928** |
| Recall | 0.851 | 0.923 | 0.862 | **0.928** |
| F1 | 0.849 | 0.922 | 0.861 | **0.927** |
| $\hat{\sigma}$(Accuracy) | 0.000 | 0.000 | 0.000 | 0.000 |

All three suggested improvements are helpful in enhancing the performance. Through the use of derived distance, the accuracy of our model on the testing data set improved significantly, from 73% accuracy (best of part (a)) to 84% accuracy (best of part (b)). Furthermore, when dedicated models were constructed by different SCTG and NAICS categories, the performance increased even further from 84.1% (best of part (a)-(b)) to 90.8% (best of part (c)-(d)) and 92.2% (best of part (e)-(f)) for NAICS and SCTG, respectively. Finally, by using stacking method to combine all models, we achieved an accuracy of 92.8% and a BA of 87.1%. Moreover, the standard errors of accuracies are also estimated by bootstrapping over the testing dataset, as displayed in the final row of each section in **Table 6**. Ensembled model (e.g., stacking or voting method) also shown the advantage of stabilizing performance over testing sets. Most ensembled models exhibited a standard deviation of accuracies that was below 0.005%. Consequently, these values were rounded down to 0.000 in **Table 6**. Note that all results in **Table 6** are run on over 1 million testing cases, and the standard deviation will be larger as the number testing samples gets lower. These results demonstrate that the model framework is reliable and suitable for use in various applications, including freight modal split analysis, traffic assignment studies, and sustainability assessments.

Other metrics to compare different predicting models were also applied. As mentioned, tree-based models provide the probabilities of selecting each mode. Thus, a threshold probability was used as a decision boundary to select a specific mode for each freight mode. By varying the decision boundary from 0 to 1 for each mode, a receiver operating characteristic curve was plotted, as depicted in **Figure 11**. The *x*-axis represents the false positive rate, and the *y*-axis represents the true positive rate. The first and second rows correspond to RF and BAG models, respectively. The third row shows the results of two ensemble methods. Each plot displays five lines that represent the receiver operating characteristic curve for each of the five modes indicated by the lower right legends. The AUC was used to evaluate model performance, where a value closer to 1 indicates better performance. The average AUC for all five plots is displayed in each plot. Using the RF or bagging method, the models built by NAICS and SCTG codes outperformed the unified ("one model for all") models. Notably, the voting and stacking models (the third row) yielded significantly better performance than applying the base learners alone (the first and second row).

The behavior of the high-performance models was examined. The left and right subplots in **Figure 12** correspond to the variable importance of the RF model and the level-2 stacking model (i.e., the meta-learner), respectively. One RF model was applied to the overall data set for **Figure 12** (left). The routed distances of different modes ("M1"–"M5" and "I1"–"I5") accounted for a significant portion of variations in the model. Conversely, a different pattern is discernible in **Figure 12** (right), where five out of top six most "expressive" variables are probabilities of choosing different modes derived from base learners/models. This demonstrates that the meta-learner, an XGBoost model, begins to make predictions with the results of base learners before further polishing predictions using other variables (e.g., "v2w," "sw"). Also, the effects of those variables were already considered by the base learners (level-1 model), as shown by **Figure 12** (left).

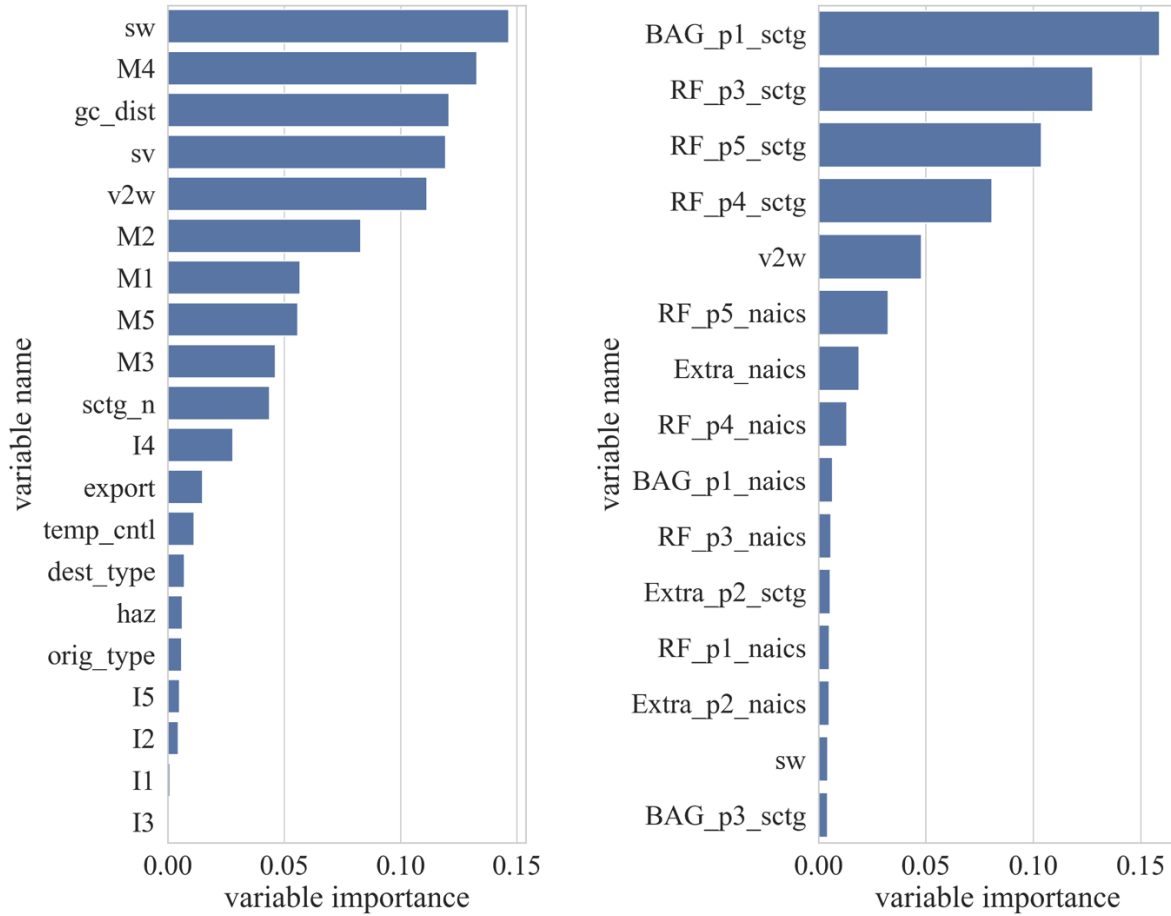

**Figure 12. Variable importance for the (left) global RF model and (right) XGBoost stacking model**

Recently, many methods have been developed for interpreting machine learning models. Among these, the LIME (Local Interpretable Model-Agnostic Explanations) approach is widely used in agnostic-model interpretation by fitting local linear additive models to approximate the model to be explained for individual inputs (Ribeiro et al., 2016). Furthermore, the SHAP (Shapely Additive Explanations) method (Lundberg et al., 2019; Uddin et al., 2023), which is based on game theory and focuses on global interpretation, was developed to evaluate the effects of different model features and has some theoretical and practical benefits. Recently, specialized algorithms are available for generating and efficiently analyzing SHAP values (Lundberg et al., 2019). Although the computational details of SHAP values can be tedious, the SHAP value only describes the "marginal impacts" of a specific feature on the final outputs. For this purpose, **Figure 13** shows a force plot for one randomly chosen sample, where the probability of selecting each mode is interpreted by the contribution of each feature. In **Figure 13** (a), value to weight ("v2w"), single weight ("sw"), and great circle distance ("gc_dist") are the top three reasons to predict the for-hire truck mode, whereas the same variables on other modes have different magnitudes of influence.

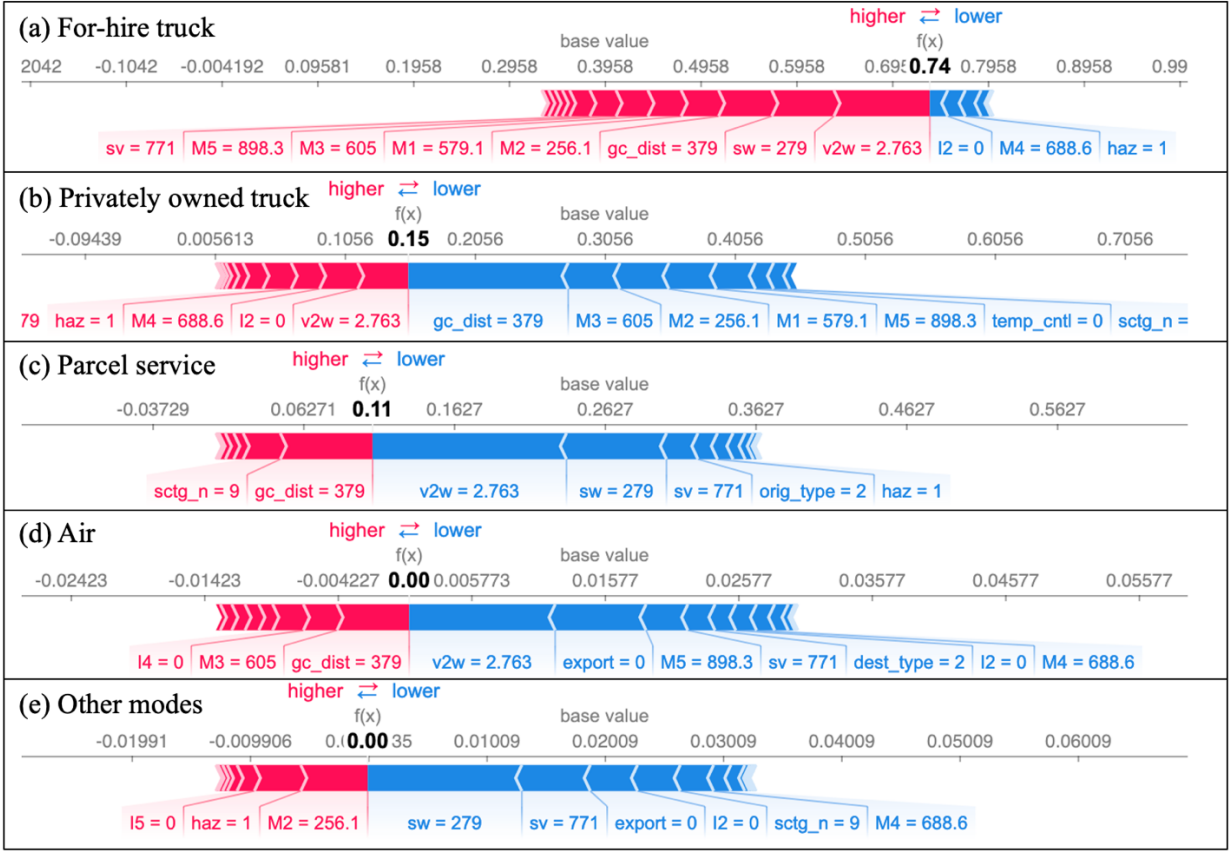

**Figure 13. A force diagram of SHAP values of one DT model over one testing data sample**

The SHAP method can also be applied to RF models, as shown in **Figure 14**. **Figure 14 (a)** displays the ratio of SHAP values across all the five modes by measuring the mean absolute SHAP values for each feature over 500 randomly selected samples. The remaining five plots (**Figure 14 (b)–(f)**) describe the SHAP values correspond to each of the five modes' output probabilities. The *x*-axis shows the SHAP values. The feature value of each dot is denoted from high to low using a red to blue scale. The features are listed by their total SHAP impacts from top to bottom. A dot with a negative effect falls at the left side of the *x*-axis, and one with a positive effect falls at the right side of the *x*-axis. As shown, different freight modes have different influential features. A single weight ("sw") had the highest SHAP value for for-hire trucks and parcel service, whereas the great circle distance ("gc_dist") had the highest SHAP value for privately owned trucks.

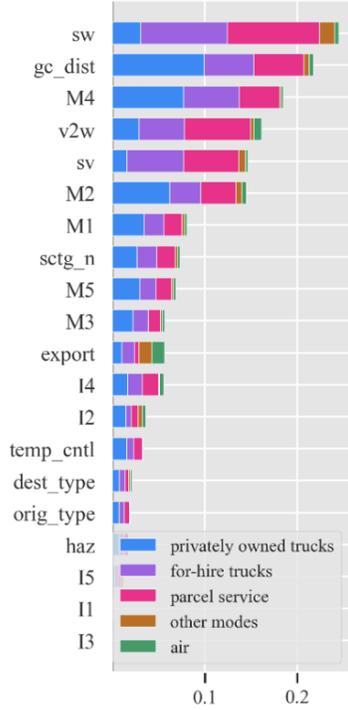
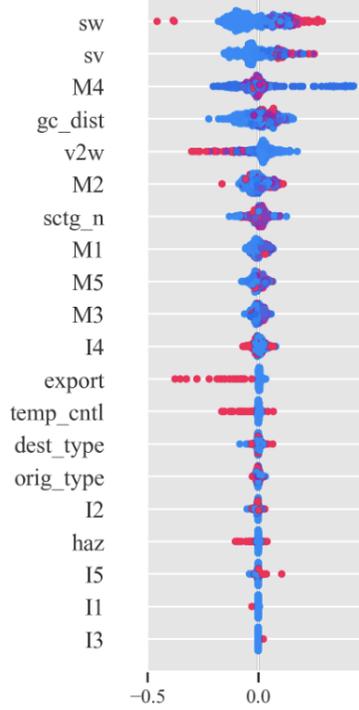
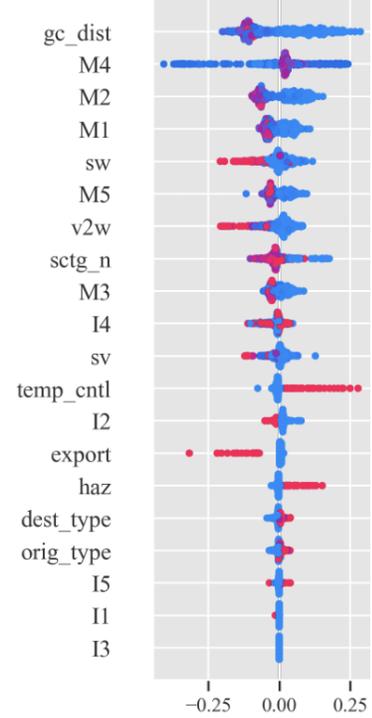
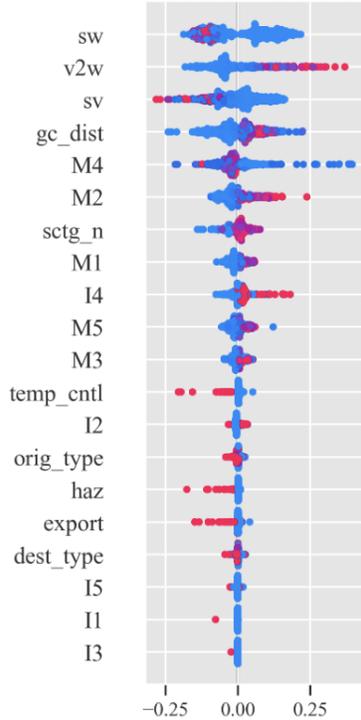
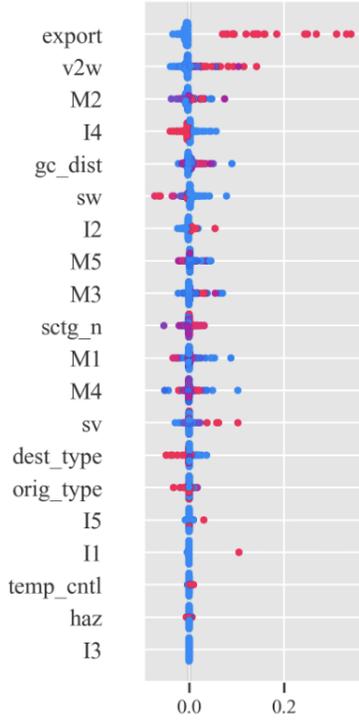
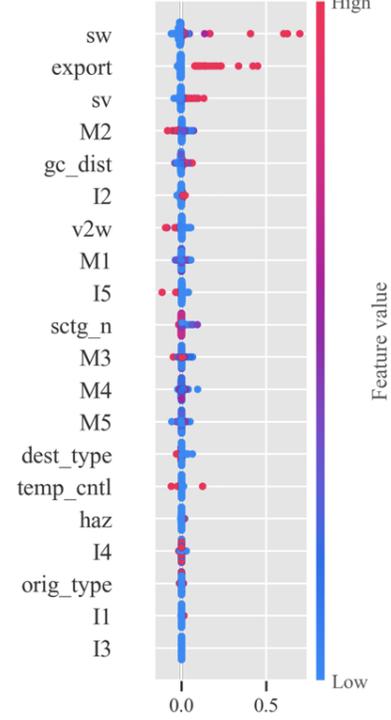

**Figure 14.** (a) Summary plot of absolute SHAP values and (b–f) swarm plots of each mode choice using the SHAP game theory approach to explain feature influences on the overall RF model

**Figure 14** reveals several nontrivial patterns that offer significant insights. For instance, exported products are more likely to be transported via air or other modes. Furthermore, the great circle distance ("gc_dist") has a strong influence on privately owned trucks' mode choice. Also, privately owned trucks rarely travel for distances greater than 500 miles in great circle distance. Notably, several features with high SHAP values for privately owned trucks include routed distances for different modes (e.g., "M4"). In contrast, single weight ("sw"), single value ("sv"), and value-to-weight ("v2w") ratios are the most influential features for for-hire trucks. This could be due to shippers operating for-hire trucks within an open market and competing with other long-distance freight modes who occasionally prefer transporting products with high single weight and single value but low value to weight. Notably, low value to weight is a relative term compared to commodities shipped via parcel service.

In addition to interpreting SHAP values on a feature-by-feature basis, we assessed the interaction values between two variables using the SHAP dependency plot, as illustrated in **Figure 15**. Each plot displays single weight ("sw") and the corresponding SHAP value. The dots denote another selected feature that has a significant interaction effect for the given freight mode. The red to blue legend indicates the feature values from highest to lowest. The results show that there is an interactive relationship between single weight ("sw") and distance metrics (e.g., "gc_dist"). For-hire truck and parcel service are sensitive to traveling distance given a fixed single weight, and privately owned trucks have reduced SHAP values for single weights when traveling long distances. Moreover, for air and other modes, the variable for indicating export shipment (0-domestic and 1-export) may have minor interaction effects on fixed single weight, as indicated by the SHAP values.

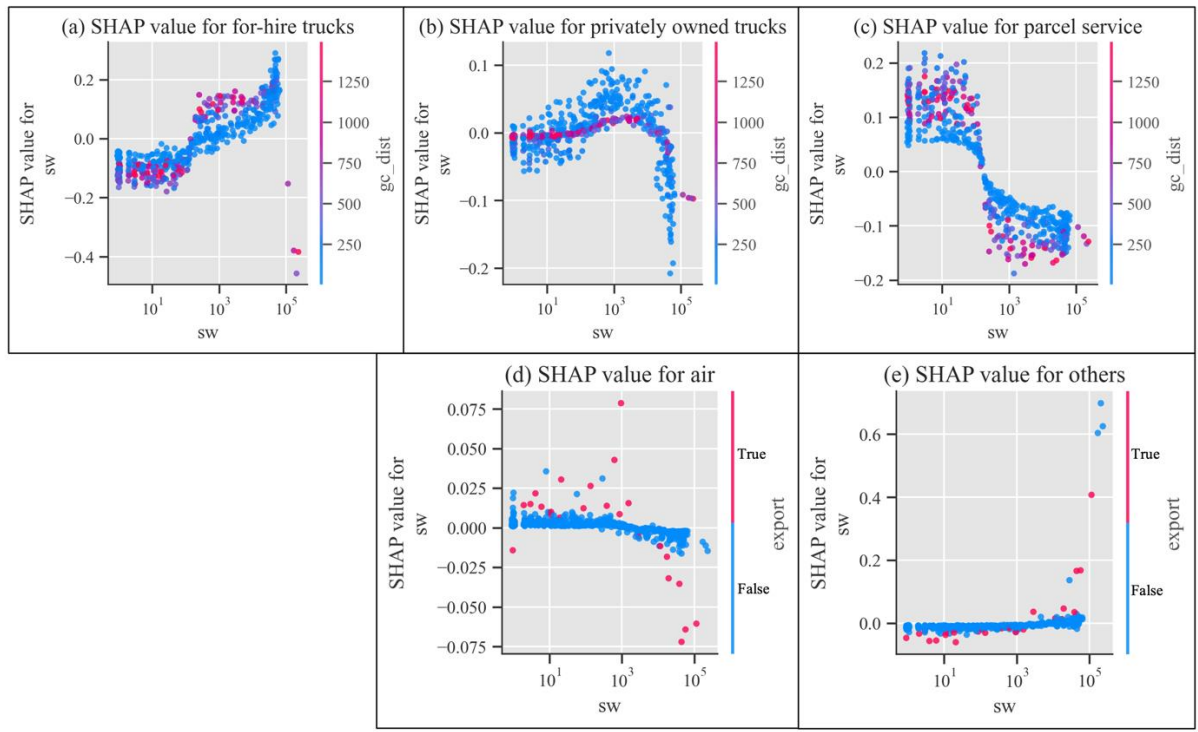

**Figure 15. SHAP dependence plot of the single weight ("sw") feature and some interaction features for each freight mode**

The previous SHAP plots illustrate the patterns of level-1 models, such as a global RF model. Analyzing the SHAP effects on level-2 of the stacking model or on concatenated models split by different commodity or industrial types can offer further insights into interpreting models. **Figure 16** presents the results of such analyses, revealing that the level-2 meta-learner simply averaged the probabilities of each model class outputted by the level-1 models. Many features are probability outputs from level-1 models, including "BAG_p1_sctg," which denotes the probabilities of selecting for-hire trucks for the "sctg" segmented bagging tree model. Furthermore, information such as single weight ("sw") and value to weight ("v2w") still contribute to model interpretation based on their SHAP values. Although binary dummy variables are used for constructing level-1 models based on SCTG and NAICS types and their effects are summed, their SHAP values do not rank among the top 10 features for all freight modes.

In summary, the proposed modeling framework and methodologies can significantly improve the performance of mode choice modeling. Besides using basic performance metrics, receiver operating characteristic can also be helpful in visualizing the model's performance. The machine learning model results were also interpreted, generating useful insights about feature importance with the help of the SHAP method.

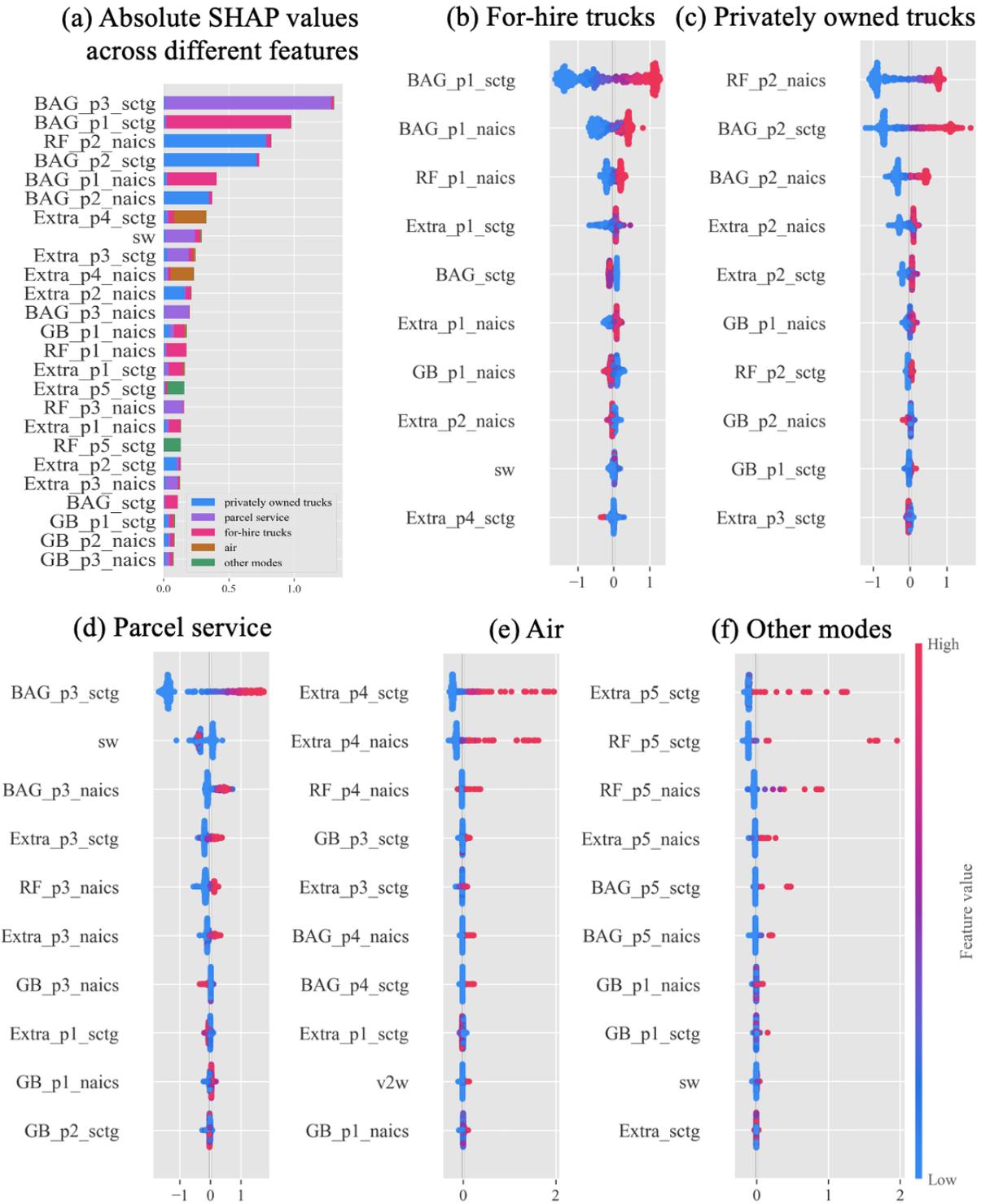

**Figure 16.** (a) Summary plot and (b–f) swarm plots of the level-2 stacking model

# 6 Conclusions

In this study, we developed a high-performance freight modal choice prediction model using the 2017 CFS PUF data set. The potential applications of such a model are wide-ranging and significant. Using some basic shipment and geographical information such as value, weight, commodity type, origin, and destination, our model predicted the probabilities of choosing available freight modes. Moreover, using ensemble techniques (e.g., stacking) for machine learning models by combining models built upon different sets of data records from different aspects, we significantly improved the model performance for freight mode prediction (or more generally, transportation mode prediction).

During the course of model development, numerous lessons were learned. First, although ensemble methods are advantageous for enhancing the overall model performance, they may not be effective when solely applied to a specific type of model (e.g., local models by SCTG). Furthermore, the performance of non-tree–based models was comparatively lower than that of tree-based models. Consequently, their inclusion in the ensemble did not lead to any significant improvement. Second, we attempted to identify different base learners from various categories dynamically to generate a dynamic model that combines different kinds of machine learning models. We hypothesized that certain commodities or industries may possess unique decision-making logic and so, a simple model could outperform tree-based models for some commodities. However, in practice, tree-based methods like RF and BAG demonstrated superior performance across all categories, including those with few data records. Finally, the use of neural networks as meta-learners in stacking methods was attempted without success.

Some limitations of this study stem from the CFS PUF data set. First, because it is a survey sampling among shippers, the data set does not provide uniform coverage of all freight movements (Comparison of Commodity Flow Survey (CFS) and Freight Analysis Framework (FAF) for 2017 Domestic Freight Flows, 2022). Another constraint of this research is the lack of consideration given to shipping time and its variability. Although logistics travel time could be estimated by reconstructing travel networks for various modes of transportation, this task is not straightforward because much of the logistics time is spent waiting at fixed locations such as warehouses, in addition to the travel time between locations. Furthermore, pricing or cost bids across different freight modes are essential factors when providing freight service to customers but are not incorporated in the data set. The only reported value is the value of a single product.

In this study, freight modes were classified into five categories to facilitate comparison with previous research. Water-related modes, train-related modes, and pipelines were all grouped under other modes because of their relatively low percentage of shipments. A nested model can be developed to further differentiate those modes. Also, one can apply oversampling methods to enhance the results for the air mode, which composed of less than 2% of trips in most commodity categories. The most challenging aspect of the study was predicting air mode shipments because of a data imbalance resulting from their small proportion in the data set. To address this issue in future work, we will focus on improving predictions for infrequent modes (e.g., air, other modes) through the use of different mode weights.

# 7 Statements and Acknowledgements

Statement: During the preparation of this work, the authors used gpt-3.5-turbo in order to improve the fluency and clarity of the writing. After using this tool, the authors reviewed and edited the content as needed and take full responsibility for the content of the publication.

Acknowledgement: The authors appreciate the funding and research opportunities provided by the Graduate Advancement Training and Education for collaborate research between the University of Tennessee, Knoxville, and the US Department of Energy's Oak Ridge National Laboratory.